%% file: main.tex
\documentclass[runningheads]{llncs}
\usepackage{graphicx}
\usepackage{booktabs}
\usepackage{tabularx}
\usepackage{hyperref}
\usepackage{paralist}
\usepackage{tabulary}
\usepackage{ragged2e}
\usepackage{wrapfig}
\usepackage{cite}
\usepackage{comment}
\usepackage{caption}
\usepackage{amsmath}
\usepackage{todonotes}
\usepackage{amssymb}
\usepackage{lipsum}
\usepackage{eurosym}
\usepackage{xfrac}

\newcommand{\mypar}[1]{\vspace{0.1em}\noindent\textbf{#1.}}

\newcolumntype{C}[1]{>{\centering\let\newline\\\arraybackslash\hspace{0pt}}m{#1}}

\newcolumntype{Y}{>{\centering\arraybackslash}X}

\begin{document}
%
\title{A Discussion on Generalization in Next-Activity Prediction}
%
%
\author{Luka Abb\inst{1}*\orcidID{0000-0002-4263-8438} \and Peter Pfeiffer\inst{2}*\orcidID{0000-0002-0224-4450}  \and Peter Fettke\inst{2}\orcidID{0000-0002-0624-4431} \and Jana-Rebecca Rehse\inst{1}\orcidID{0000-0001-5707-6944}}
\authorrunning{Abb et al.}
%
 \institute{University of Mannheim, Germany \\
 \email{\{luka.abb, rehse\}@uni-mannheim.de},
 \and
 German Research Center for Artificial Intelligence (DFKI) and Saarland University,
Saarbrücken, Germany \\
 \email{\{peter.pfeiffer, peter.fettke\}@dfki.de}}
\maketitle              
\def\thefootnote{*}\footnotetext{Equal contribution}\def\thefootnote{\arabic{footnote}}
\begin{abstract}
The goal of next-activity prediction is to forecast the future behavior of running process instances. Recent publications in this field predominantly employ deep learning techniques and evaluate their prediction performance using publicly available event logs. This paper presents empirical evidence that calls into question the effectiveness of these current evaluation approaches. We show that there is an enormous amount of example leakage in all of the commonly used event logs and demonstrate that the next-activity prediction task in these logs is a rather trivial one that can be solved by a naive baseline. We further argue that designing robust evaluations requires a more profound conceptual engagement with the topic of next-activity prediction, and specifically with the notion of generalization to new data. To this end, we present various prediction scenarios that necessitate different types of generalization to guide future research in this field.

\end{abstract}

\section{Introduction}
Predictive process monitoring (PPM), or process prediction, is a branch of process mining that is concerned with the forecasting of how a running process instance will unfold in the future \cite{PM_Handbook}. For example, PPM approaches may predict what the outcome of the process instance will be, how long it will take to complete, or which activities will be executed next. In contrast to techniques like process discovery or conformance checking, process prediction is forward-facing, and aims to identify process execution problems like delays or compliance violations \emph{before} they occur, thus enabling an organization to preemptively take preventive counteractions \cite{PM_Handbook}. 

Whereas older approaches to process prediction relied on explicit models of process behavior, such as transition systems 
or probabilistic automata \cite{breuker_2016_comprehensible}, recent research has almost exclusively tackled the problem with neural networks~\cite{Evermann2017}. 
The majority of this research has also focused on control-flow predictions, specifically the prediction of the \emph{next activity} in a trace~\cite{Neu2022}. 
At a high level, all existing contributions approach next activity prediction as a self-supervised machine learning problem~\cite{rama2021deep,mida,MPPN}: An existing event log is randomly split into a training and a test set. 
A machine learning model, typically a deep neural network, is shown incomplete traces from the training set, such that it learns to predict the next activity in that trace. 
The performance of the trained model is then evaluated by predicting the next activity for incomplete traces of the unseen test set and computing performance measures. 
Almost all existing publications train and evaluate their models on a relatively small collection of event logs for their evaluation. This includes the Helpdesk event log \cite{helpdesk} and the logs from the Business Process Intelligence Challenges (BPIC) 2012, 2013, and/or 2017. 

In this paper, we argue that this current way of training and evaluating next activity prediction models is biased in the sense that it does not evaluate how well these models would generalize to unseen data. We argue that, in order to design reliable evaluation procedures, it is necessary to first engage with the topic of next-activity prediction on a more conceptual level. 
Our line of argument is based on several observations about the aforementioned event logs: 
First, the next-activity label is almost entirely determined by the control-flow of the prefix.
Second, when only considering the control-flow perspective, there is an enormous amount of example leakage in all logs, so that most predictions are made on prefixes that were already seen during training. Third, as other research has already shown \cite{pfeiffer2023label}, incomplete traces can often continue in different ways, so that the maximal achievable accuracy in this evaluation setting is unknown and probably much lower than 100\%.

After introducing basic concepts in \autoref{sec:background}, we provide empirical evidence for each of these observations and demonstrate that the next-activity prediction task in these event logs is a rather trivial one that can be solved by a naive baseline (\autoref{sec:validity}). 
\autoref{sec:generalization} presents various scenarios for generalization in process prediction which are grouped into three types of generalization. Finally, we discuss related work in \autoref{sec:related_work} and conclude the paper in \autoref{sec:conclusion}

\section{Background} \label{sec:background}

\mypar{Event Log Data}
PPM works on \emph{event log data}, gathered from the execution of business processes in information systems. 
An event log is a collection of cases. A case is represented by a \emph{trace} $t$, i.e., a sequence of events $\langle e_{1},\dots,e_{n}\rangle$ of length $n$. 
Each event $e$ has two mandatory attributes: the \emph{activity} and the \emph{case ID}. 
In addition, events can have additional attributes, such as a timestamp or an executing resource, which describe the context in which the event has occurred. 
Similar to events, traces can also have additional attributes, such as an allocated project. 
A case represents a completed process execution. 
For PPM, we are interested in predicting the future behavior of running cases, which are represented by trace prefixes. A \emph{trace prefix} of a trace $t$ of length $p$ is defined as a subsequence $\langle e_{1}, \dots ,e_{p}\rangle$, with $1 \leq p < n$.

\mypar{Next Activity Prediction}
The goal of next activity prediction is to predict which activity is performed next in a running case. 
Formally, this problem is framed as multi-class classification, where each class represents one activity.  
For each trace $t$ in a given event log, pairs $(x,y)$ of features $x$ and labels $y$ are created. $x$ is a prefix of $t$ with length $p$, which represents the running case. $y$, which is often called the label of $x$, represents the activity at position $p+1$ of $t$, i.e., the next activity, which should be predicted. 
These pairs $(x,y)$ are provided to a machine learning model, typically a deep neural network, such that it learns a predictor function $f$ that maps the prefix to the correct next activity, i.e., the class to which the prefix belongs. 
To learn and evaluate $f$, the event log is split into two parts, the training set and the test set. 
The model is trained on the prefix-label pairs from the training set and evaluated on those from the test set. Therefore, for each prefix $x$, its prediction $\hat{y} := f (x)$ is compared with the ground truth label $y$ and performance measures like accuracy and F1 score can be computed.

\section{Validity Issues in Existing Research} \label{sec:validity}
In this section, we examine various phenomena that pose threats to the validity of next-activity prediction research. To substantiate our discussion, we present empirical evidence that was generated in a setting that is representative of the typical evaluation setup used in the field. We employ five commonly used event logs (Helpdesk, BPIC12, BPIC13 Incidents, BPIC17 Offer, and MobIS \cite{Mobis}) and generate six splits for each log: five in which we randomly allocate traces so that 80\% of them are part of the training set and 20\% are part of the test set, and one in which the split is time-based so that the 20\% of traces with the most recent start timestamps end up in the test set. We then generate $n-1$ prefix-label pairs $(x,y)$ from each trace with lengths $p \in [1, n-1]$ and calculate prediction accuracy as the percentage of prefixes in the test set for which the correct next-activity label was predicted, i.e., $\hat{y}=y$. We do not apply log preprocessing or make any other changes to the data. The code and data needed to reproduce our results are available at \url{https://gitlab.uni-mannheim.de/jpmac/ppm-generalization}.


\subsection{Example Leakage}
Leakage in machine learning refers to information being made available to a model during training that it would not have access to when classifying unseen data \cite{leakage}. This can lead to an unrealistic assessment of the model's performance with respect to the classification task at hand. One particular type of leakage is \emph{example leakage}, which occurs when the same example (more specifically, the same feature vector) is present in both the training and the test set. In this case, the classification is a trivial one, as the model is not required to learn generally-valid relationships between features and labels. Example leakage can be a considerable problem when doing prediction on event logs, due to the repetitive nature of the process executions recorded in them \cite{leakage2}.  

In order to quantify example leakage in next-activity prediction, we first need to establish when two prefixes can be considered identical. 
We can limit the set of features that need to be considered when establishing equality to those that are actually relevant for predicting the next activity. Previous research has already examined the extent to which context attributes, such as resource or time, enhance prediction performance compared to solely considering the previous control-flow recorded in a prefix \cite{Brunk2020}. They have found that, in most cases, including context attributes does improve predictions compared to only considering control-flow features, but that these improvements are rather insignificant (low single-digit percentage increases in accuracy). Based on these findings, we can conclude that in most cases, the next-activity label can be correctly predicted when only the control-flow of the prefix is known. In the following, we therefore consider two prefixes to be identical if they exhibit the same control-flow, i.e., if they have the same activities in the exact same order. 

\begin{figure}[htb]
    \centering
    \includegraphics[width=0.75\linewidth]{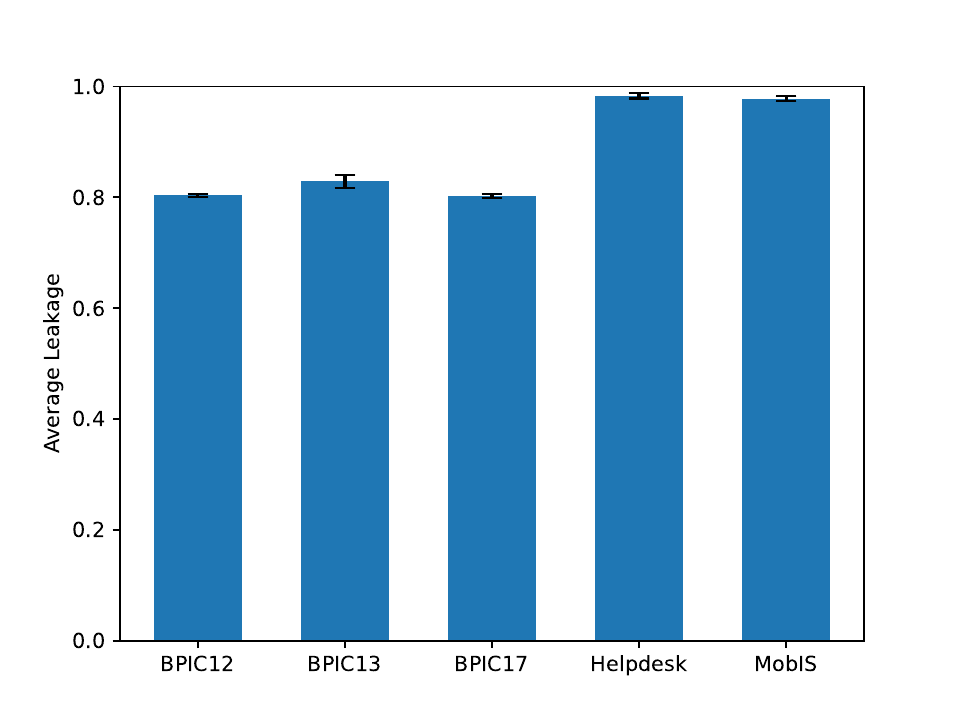}
    \caption{Example leakage percentage for each event log, averaged over the splits.}
    \label{fig:leakage}
\end{figure}

With this equality criterion, we can now quantify example leakage by calculating the percentage of prefixes in the test set that is also included in the training set.
The amount of example leakage in the event logs commonly used for the evaluation of next-activity prediction techniques is shown in (\autoref{fig:leakage}). We observe that, across all datasets and splits, example leakage is above 80\%, and even close to 100\% in the Helpdesk and MobIS event logs. This means that most of the predictions made on the test set are trivial ones, and consequently, that one cannot draw valid conclusions about how well a prediction model would perform on unseen data from this evaluation setting.

\subsection{Baseline and Accuracy Limit}
We can further illustrate this issue by demonstrating that the prediction accuracy of state-of-the-art models lies in a relatively narrow corridor that is bounded by a naive baseline with little to no generalization capacity on the lower end, and by the maximal accuracy that can be achieved with only control-flow features on the upper end.

We construct the baseline as follows: for each unique prefix in the training set $x:=\langle e_{1}, \dots ,e_{p}\rangle$, where $e$ represents the activity only, it simply predicts the most common next activity. If an unknown prefix is encountered (i.e., an example that has not leaked), it instead predicts the most common next activity associated with only the last activity $e_p$ in the prefix, similar to a bigram model, i.e., $x:=e_{p}$.

The upper bound is based on the observation that a common implicit assumption in supervised learning, that each unique combination of feature values maps to exactly one label, does not hold in the process mining domain. Event logs nearly always contain traces that have identical control-flow up to a point but diverge afterwards, for example due to exclusive continuation paths or concurrent activity execution. In the context of next-activity prediction, this means that a prefix exhibits \emph{label ambiguity} \cite{pfeiffer2023label}.
If a prediction model that predicts a single next-activity label is tasked with classifying a label-ambiguous prefix, the best prediction in terms of the resulting overall accuracy it can make is the activity that is most frequently associated with that prefix. All other activities will never be predicted.

From this, we can derive that there is an \emph{accuracy limit} that a prediction model can achieve on a given (test) dataset when it only makes predictions based on the control-flow of the prefix. This accuracy limit is simply calculated as the percentage of examples in the test set in which the label is the most common label for the corresponding prefix.

\begin{figure}[htb]
    \centering
    \includegraphics[width=1.\linewidth]{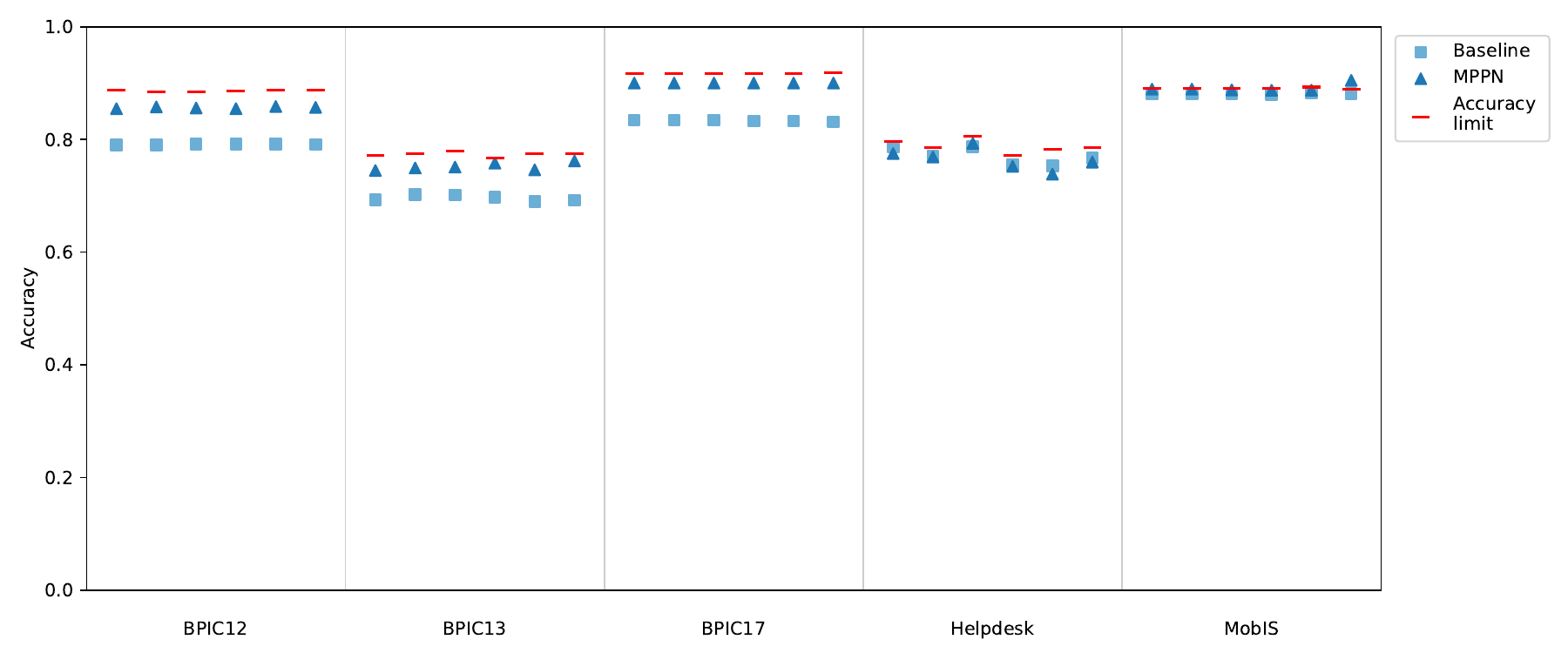}
    \caption{Prediction accuracy of the naive baseline and the MPPN neural network, along with the accuracy limit in the test set. Each split plotted separately.}
    \label{fig:benchmark}
\end{figure}

\autoref{fig:benchmark} shows the prediction accuracy achieved by the baseline prediction model described above and the MPPN \cite{MPPN}, a state-of-the-art neural network predictor that includes contextual attributes for its prediction. The accuracy limit for each test split is also included. Of course, this comparison is limited since it only includes a single state-of-the-art model. However, given that benchmark experiments in previous research have consistently shown that many next-activity prediction models achieve almost the same accuracy when evaluated on the same data (e.g., \cite{rama2021deep,mida,MPPN}), our observations are likely to apply to other models as well. 

In the Helpdesk and MobIS logs, the training and test set almost completely overlap. Predicting the next activity in these event logs is therefore trivial, and consequently, both models achieve the same prediction accuracy. In fact, the only reason that they do not reach 100\% accuracy is label ambiguity, which is why the observed accuracy for these models is almost identical to the accuracy limit.
In other event logs, which exhibit slightly less example leakage, the accuracy of the naive baseline is still very close to the one of the state-of-the-art model, although there is a notable gap of a few percentage points. It is, however, unclear to which extent this performance gap can be attributed to the MPPN's ability to generalize to unseen examples. An alternative explanation would be that its consideration of context features allows it to resolve the label ambiguity in some traces, and thereby improve its predictions, whereas the baseline only considers control-flow features; this would be consistent with the findings of \cite{Brunk2020}, i.e., that incorporating context slightly improves prediction accuracy. 

Given that the evaluation setting that we used in this section has been so widely employed in existing publications on next-activity prediction, our findings suggest that a significant portion of the perceived advancements in the field may be -- in a sense -- illusory. As a research community, we now have a large number of proposed next-activity prediction techniques that employ several different neural network architectures, inductive biases, and strategies to incorporate different types of features. However, we have very little idea to what extent these techniques would be able to generalize well enough to make good predictions on unseen data -- and consequently, if they would be able to provide value in a real-world application.

Although it would also be possible to address the issues that we have pointed out in this section on a technical level, we argue that they are symptomatic of a broader problem in process prediction research, namely a lack of engagement with the topic on a conceptual level. In particular, we believe that there is an insufficient understanding of what \emph{generalization} means in a process prediction context.


\input{generalisation}

\section{Related Work} \label{sec:related_work}
So far, the conceptual flaws of process prediction beside label ambiguity \cite{pfeiffer2023label} have been discussed little. The majority of papers have introduced new approaches for process prediction, starting from the first deep learning based model \cite{Evermann2017}, to more complex architectures \cite{mida,MPPN}.
Limited work has been conducted to ensure realistic evaluation settings or test generalization capabilities.

Weytjens et al. \cite{leakage2} introduce a pre-processing algorithms to prevent leakage in process prediction focusing on the remaining time prediction problem. Their approach splits the traces on a temporal basis such that there is no temporal overlap between the prefixes used for training and test. However, this does not prevent example leakage on prefix-level.

In \cite{Tax2020Nov}, the authors compare discovery-based algorithms to sequence-learning algorithms in terms of their accurateness and generalization capabilities. The event logs are split into training and test sets. However, as the paper does not mention any technique to prevent example leakage, it is very likely that the splits used in the experiments face a similar high portion of leaked prefixes which limits the validity of generalization capabilities measured.

Peeperkorn et al. \cite{Peeperkorn2022Dec} propose an evaluation strategy to leave certain variants out of the training set and only have them in the test set. They used this splitting strategy to evaluate whether prediction models can learn process model structure of the unknown system behind the log, focusing mainly on concurrent activities in process models. Thus, they did not systematically cover all generalization scenarios introduced in this paper. They found that the generalization capabilities of LSTM prediction models are inversely correlated with the number of variants left out. However, as they measured with accuracy, it is unclear how label ambiguity affected the experiments. In comparison to their work, we propose several generalization scenarios.


\section{Conclusion} \label{sec:conclusion}


In this paper, we have critically analyzed the current procedure of evaluating PPM algorithms in research and found that little to no generalization capabilities can be tested that way.
The proposed generalization scenarios can be used to measure how much difference between train and test set there is and which generalization capabilities are required for which log, i.e., which scenarios are present and which not. Furthermore, synthetic event logs containing these pattern can be simulated and existing one split accordingly to test for generalization. Guided by the plausible predictions, new prediction algorithms can be developed that specifically account for these.

While the generalization scenarios are inspired by real-world situations, real event logs are required for setting the ground truth label of unseen prefixes. In the scenarios presented, we assumed a ground truth label and argued whether such a prediction will show generalization. In some scenarios, the expected label is more clear than in other scenarios. However, these are only plausible predictions. Real generalization can only be tested if the ground truth label is not assumed but determined by the data.

Although we have focused on next-activity prediction and other prediction situations were out of scope for this work, there might be more scenarios in next-activity prediction that are not yet covered. Furthermore, the high percentage of example leakage between train and test set raises the question whether generalization capabilities are actually required if the behaviour in both sets is that similar when considering the control-flow only. Following that, prediction models that take context information into account might actually be able to generalize with respect to the scenarios of unseen attribute value combination, as they reach comparable or higher accuracy as control-flow only models. Nevertheless, this has yet not been shown explicitly.

In the future we plan to create a benchmark set of event logs that cover the presented generalization scenarios.
Furthermore, the scenarios can be adopted to other prediction tasks like outcome prediction.



\vspace{-0.5em}
\bibliographystyle{splncs04}
\bibliography{bibliography}

\end{document}

%% file: generalisation.tex
\section{Generalization in Process Prediction} \label{sec:generalization}
In machine learning, generalization refers to the ability of a trained model to make correct predictions on samples that it has not seen during training. 
This in an important capability because a model should not only be able to handle the samples that it is already familiar with, but also other samples that it will be faced with when applied in its respective application context. 

As pointed out in the previous section, splitting an event log into training and test sets with the goal of having samples in the test set that were not present in the training set does not work as expected. 
This means that, although generalization is a characteristic of interest for machine learning in general and process prediction in particular, the generalization capabilities of PPM algorithms have so far not been explicitly evaluated, in the sense of applying an algorithm on a test log that has little to no overlap with the training data\footnote{A notable exception to this is \cite{Peeperkorn2022Dec}, which focuses on process model structures}. 
Such an evaluation is undoubtedly necessary, but it requires a discussion on what generalization means in a process context and how it should be measured. 

To contribute to this discussion, this section presents several exemplary prediction scenarios, classified into different generalization types, and discusses which predictions a PPM algorithm should reasonable make in each. These scenarios are not meant to be complete. Rather, they are intended to serve as a starting point for understanding generalization in process prediction.

\subsection{Prediction Scenarios}

In all scenarios, we suppose to train a prediction model on the mentioned log, i.e., we create all prefixes for all traces $t$ in the log $L$ and train the model on the resulting samples $(x, y)$. For each scenarios we show prefixes that are not seen so far, i.e., that are not included in the log. Given the unseen prefix as input to the model, we explain which predictions are plausible to be made. 
Thus, we only assume what could be the correct ground truth label. 
If the model is able to make this prediction on the unseen prefix, we say that it can generalize in this scenario. 

In all scenarios, we focus on the problem of predicting the next activity only. Predicting attributes like resource, time or properties like the process outcome are related problems, but the correct predictions differ, so they require a separate discussion. Furthermore, we assume that we do not have access to additional information like a process model; only the observations in the event log are given. 

\begin{table}[h]
    \parbox{.29\linewidth}{
        \centering
        \begin{tabular}{| C{4cm} |}
            \hline
             Event Log $L1$                                          \\ \hline \hline 
             $\langle \text{A, B, C1, C2, C3, D, E} \rangle$        \\ 
             $\langle \text{A, B, C2, C1, C3, D, E} \rangle$        \\
             $\langle \text{A, B, C2, C3, C1, D, E} \rangle$        \\
             $\langle \text{A, B, C3, C1, C2, D, E} \rangle$        \\
             $\langle \text{A, B, C3, C2, C1, D, E} \rangle$        \\ \hline
        \end{tabular}
        \caption{Concurrency}
        \label{tab:sample_log_concurrency}
    }
    \hfill
    \parbox{.29\linewidth}{
        \centering
        \begin{tabular}{| C{4cm} |}
            \hline
             Event Log $L2$                                          \\ \hline \hline 
             $\langle \text{A, B, C, D, E, F, \underline{G, H}} \rangle$        \\ 
             $\langle \text{A, B, C, F, D, \underline{G, E}, H} \rangle$        \\
             $\langle \text{A, B, C, D, F, E, \underline{G, H}} \rangle$        \\
             $\langle \text{A, B, F, C, D, \underline{G, H}, E} \rangle$        \\ \hline
        \end{tabular}
        \caption{Concurrency with label ambiguity}
        \label{tab:sample_log_concurrency_ambiguity}
    }
    \hfill
    \parbox{.22\linewidth}{
        \vspace{1.1cm}
        \centering
        \begin{tabular}{| C{3cm} |}
            \hline
             Event Log $L3$                                          \\ \hline \hline
             $\langle \text{A, B, C, D} \rangle$        \\ 
             $\langle \text{A, B, B, C, D} \rangle$        \\ \hline
        \end{tabular}
        \caption{Loops}
        \label{tab:sample_log_loop}
    }
    \vspace{-2em}
\end{table}

\subsubsection{Unseen Control-flow} \label{sec:concurrency}
Log $L1$ in \autoref{tab:sample_log_concurrency} shows the scenarios where activities $C1$, $C2$ and $C3$ can occur in any order. $L2$ in \autoref{tab:sample_log_concurrency_ambiguity} shows a similar, yet more complex scenario with $C$, $D$, $E$, $F$, $G$, $H$ in any order. This can be caused, e.g., by concurrent activities and is a common phenomenon in real-world event logs. Another common scenario is the appearance of activities that can be executed multiple times after another as shown in $L3$. For event logs with such patterns, four interesting scenarios can occur:

\begin{enumerate}
    \item $L1$ and prefix $\langle \text{A, B, C1, C3, C2, D} \rangle$. Expected prediction: $E$. Although the model has not seen this prefix due to a new order of $C1$, $C2$ and $C3$, it should have learned that the case always continues with $E$ after $D$, regardless of the order of the previous activities.
    \item $L1$ and prefix: $\langle \text{A, B, C1, C3, C2} \rangle$. Expected prediction: $D$. Again, the prediction model should have learned that regardless of the order of $C1$, $C2$ and $C3$, $D$ always follows.
    \item $L2$ and prefix: $\langle \text{A, B, C, D, F, G} \rangle$. As seen in $L2$, both $E$ and $H$ have happened after $G$. However, in each trace, either $E$ or $D$ directly follows $G$. This is the situation of label ambiguity described in \cite{pfeiffer2023label}. Both options, $E$ and $D$ are valid continuations and thus valid predictions. 
    \item $L3$ and prefix: $\langle \text{A, B, B, B, C} \rangle$. Expected prediction: $D$. The model should have learned that the case always continues with $D$ after $C$, no matter how often $B$ has happened.
\end{enumerate}

\subsubsection{Unseen Attribute Value Combinations} \label{sec:unk_context}
\begin{table}[h]
    \parbox{.45\linewidth}{
        \centering
        \begin{tabular}{| C{6cm} |}
            \hline
             Event Log $L4$                                          \\ \hline \hline
             $\langle \text{(\textbf{A}, R1), (\textbf{B}, R100), (\textbf{C}, R2)} \rangle$        \\ 
             $\langle \text{(\textbf{A}, R1), (\textbf{B}, R101), (\textbf{C}, R2)} \rangle$        \\
             $\langle \text{(\textbf{A}, R1), (\textbf{B}, R101), (\textbf{C}, R2)} \rangle$        \\ \hline
        \end{tabular}
        \caption{Example Log with different resources $R$ performing $B$}
        \label{tab:sample_context_R}
    }
    \hfill
    \parbox{.45\linewidth}{
        \centering
        \begin{tabular}{| C{6cm} |}
            \hline
             Event Log $L5$                                          \\ \hline \hline
             $\langle \text{(\textbf{A}, 2\euro), (\textbf{B}, 2\euro), (\textbf{C}, 2\euro)} \rangle$        \\ 
             $\langle \text{(\textbf{A}, 499\euro), (\textbf{B}, 499\euro), (\textbf{C}, 499\euro)} \rangle$        \\ 
             $\langle \text{(\textbf{A}, 501\euro), (\textbf{B}, 501\euro), (\textbf{D}, 501\euro)} \rangle$        \\ \hline
        \end{tabular}
        \caption{Example Log with decision depending on cost after $B$.}
        \label{tab:sample_context_C}
    }
    \vspace{-3em}
\end{table}

\begin{table}[]
    \centering
        \begin{tabular}{| C{8cm} |}
            \hline
             Event Log $L6$                                          \\ \hline \hline
             $\langle \text{(\textbf{A}, May 2022), (\textbf{B}, June 2022), (\textbf{C}, June 2022)} \rangle$        \\ 
             $\langle \text{(\textbf{A}, July 2022), (\textbf{B}, July 2022), (\textbf{C}, July 2022)} \rangle$        \\
             $\langle \text{(\textbf{A}, April 2023), (\textbf{B}, May 2023), (\textbf{D}, May 2023)} \rangle$        \\ \hline
        \end{tabular}
        \caption{Example Log with concept drift in 2023.}
        \label{tab:sample_context_T}
        \vspace{-2em}
\end{table}
In certain scenarios, the context attributes like involved resources, timestamp or cost carry important information to determine the continuation of the process instance \cite{Mobis, Brunk2020}.
Considering the contextual information is an important capability when dealing with event logs which distinguishes next step prediction from other sequential prediction tasks. As an example, we show three scenarios where we expect the prediction model to generalize in presence of context attributes. Note that in these scenarios, the models have seen the context attribute values before, i.e., they are not completely new. Just the combination of activity and context has not been seen so far.
The first example, $L4$ in \autoref{tab:sample_context_R}, shows a situation in which different resources are involved in the activities. Log $L5$ in \autoref{tab:sample_context_C} gives an example where the next activity to execute depends on the amount of Euro. Lastly, log $L6$ in \autoref{tab:sample_context_T} shows an example where timestamps are involved. 


\begin{enumerate}
    \item $L4$ and prefix $\langle \text{(\textbf{A}, R1), (\textbf{B}, R1)} \rangle$. Expected prediction: $C$. In $L4$, different resources are involved in activity $B$. However, $C$ follows $B$ every time. Thus, the prediction model should know that regardless of the resource $R$ in activity $B$, $C$ always follows.
    \item $L5$ and prefix $\langle \text{(\textbf{A}, 2\euro), (\textbf{B}, 499\euro)} \rangle$. Expected prediction: $C$. The value of Euro has changed to $499\text{\euro}$. However, the model should have learned that with $499\text{\euro}$ $C$ still follows.
    \item $L6$ and prefix: $\langle \text{(\textbf{A}, July 2022), (\textbf{B}, May 2023)} \rangle$. Expected prediction: $D$. In 2023, a drift happened causing activity $D$ to follow $B$ instead of $C$, which the prediction model should be able to express.
\end{enumerate}

\subsubsection{Unseen Attribute Values} \label{sec:new}
Sometimes, the training log might not be complete with respect to the activities or other attributes contained. For instance, a new activity (e.g. due to new requirements in the process) or a new resource (e.g. a new person joining the process/company) might occur. To demonstrate these scenarios, we use the logs $L4$, $L5$ and $L6$ from the previous section but discuss other prefixes.

\begin{enumerate}
    \item $L4$ and prefix $\langle \text{(A, R1), (F, R100)} \rangle$. As $F$ is an activity the prediction model has never seen before, there is no evidence from the event log how to continue. One option is to indicate that the model does not know, e.g., by predicting a special $\mathit{UNKNOWN}$ token. Another option would be to predict any label from the event log that could follow potentially, e.g., $C$ as this has happened in the third position in all traces in the log. 
    \item $L4$ and prefix $\langle \text{(A, R1), (B, R37)} \rangle$. This scenario is similar to the previous one but with resource $R37$ never seen before. Again, the model could indicate that it does not know or predict any label on positional basis, e.g., $C$.
    \item $L5$ and prefix $\langle \text{(\textbf{A}, 200\euro), (\textbf{B}, 200\euro)} \rangle$. The value $200\text{\euro}$ is between the seen values $2$\text{\euro} and $499$\text{\euro}. Thus, we argue that the prediction model should predict $C$.
    \item $L6$ and prefix $\langle \text{(\textbf{A}, June 2024), (\textbf{B}, June 2024)} \rangle$. The model should know that the process has changed in 2023. If tasked with 2024, the most probable next activity is $D$.
\end{enumerate}

\subsection{Implications}
Generalization over unseen control-flow constructs involves dealing with unseen control-flow variants in the prefix as shown in the scenarios in event logs $L1$, $L2$ and $L3$ in \autoref{tab:sample_log_concurrency}, \autoref{tab:sample_log_concurrency_ambiguity} and \autoref{tab:sample_log_loop}.
We assume that all activities in prefix and label are known but the specific prefix has not been seen so far. 
The event log $L2$ in \autoref{tab:sample_log_concurrency_ambiguity} is a special scenario as it is linked to label ambiguity \cite{pfeiffer2023label}. Both options $E$ and $H$ are valid prediction. However, a deterministic model will always make the same prediction when tasked with the same prefix. As $H$ has the higher frequency, the prediction model will most likely always predict $H$ although it should have -- and probably has -- learned that $E$ can also follow. When evaluating process prediction methods with point-measures like top-1 accuracy, which consider only the single most probable prediction, one cannot assess generalization properly as it does not take into account whether the model has learned that more than one option can follow. When evaluating whether the model has learned that more than one option can follow, probabilistic measures can be used that assess how much probability is given for each option.

For generalization over unseen context combinations, the prediction model must be able to interpret the context attributes and to distinguish between those scenarios where the context attributes have influence on the next activity to be predicted and those scenarios where they do not. This involves scenario as shown in logs $L4$, $L5$ and $L6$ in \autoref{tab:sample_context_R}, \autoref{tab:sample_context_C} and \autoref{tab:sample_context_T}. There can be much more complex scenarios with other context attributes where the next activity depends not only on one but the combination of multiple attribute values. 

Generalization over new and unknown attribute values are scenarios where a new attribute value like a completely unknown activity or resource occur. In such scenarios, defining plausible predictions is often not trivial and might depend one the use-case. Furthermore, dealing with never seen attribute values in the input is challenging as the model has to have learned whether there is a influence on the process or not - and in case there is which influence it has. For numerical and temporal attributes, unseen attribute values are more diverse. For instance, the number of unique values for cost in \autoref{tab:sample_context_C} can be very large and the chance that all values have been seen is rather low. Similarly, temporal attributes can be continuous and the prediction model might in practice be tasked with prefixes with year 2024 or 2025. The most reasonable approach is to make a decently confident prediction for the most likely next activity and to indicate whether the model knows the correct answer or whether it does not know. For instance, the model might predict a certain activity which usually occurred in this position in the trace but at the same time indicate that it did predict this activity only on positional basis as it has never seen this attribute value in the trace.

In practice, these scenarios might not occur in isolation. For instance, an unseen sequence of activities in the prefix can also come with unseen combination of context attributes or new attribute values which makes generalization in process prediction a challenging task.


